\documentclass[12pt]{colt2016}

\usepackage{times}

\usepackage{algorithm}
\usepackage{algorithmic}

\usepackage{tabulary}
\usepackage{booktabs}

\usepackage{epsf,psfrag,amsfonts, xcolor,array}
\usepackage{amsmath}
\usepackage{multirow}
\usepackage{cases}
\usepackage{verbatim}
\usepackage{xspace}
\usepackage{bbm}
\usepackage{mathrsfs}
\usepackage{setspace}
\usepackage[fleqn,tbtags]{mathtools}
\usepackage{tikz}

\usetikzlibrary{positioning}

\usepackage[colorinlistoftodos, textwidth=20mm]{todonotes}
\usepackage{color}

\renewcommand{\Re}{\mathbb{R}}

\newcommand{\A}{\mathcal A}

\newcommand{\X}{\mathcal X}

\newcommand{\Y}{\mathcal Y}

\newcommand{\Prob}{\mathbb P}

\newcommand{\R}{\mathcal{R}}

\newcommand{\wb}{\overline}

\newtheorem*{theorem*}{Theorem}

\title[Open Problem: Approximate Planning of POMDPs in the class of Memoryless Policies]{Open Problem: Approximate Planning of POMDPs in the class of Memoryless Policies}

 \coltauthor{\Name{Kamyar Azizzadenesheli} \Email{kazizzad@uci.edu}\\
 \addr University of California, Irvine
 \AND
 \Name{Alessandro Lazaric} \Email{alessandro.lazaric@inria.fr}\\
 \addr French Institute for Research in Computer Science and Automation (Inria)
  \AND
 \Name{Animashree Anandkumar} \Email{a.anandkumar@uci.edu}\\
 \addr University of California, Irvine
 }

\usepackage{hyperref}

\usepackage[explicit]{titlesec}
\titlespacing{\paragraph}{0pt}{0.5em}{0.5em}[]

\begin{document}
\large

%
\maketitle


\begin{abstract}
Planning plays an important role in the broad class of decision theory. Planning has drawn much attention in recent work in the robotics and sequential decision making areas. Recently, Reinforcement Learning (RL), as an agent-environment interaction problem, has brought further attention to planning methods. Generally in RL, one can assume a generative model, e.g. graphical models, for the environment, and then the task for the RL agent is to learn the model parameters and find the optimal strategy based on these learnt parameters. Based on environment behavior, the agent can assume various types of generative models, e.g. Multi Armed Bandit for a static environment, or Markov Decision Process (MDP) for a dynamic environment. The advantage of these popular models is their simplicity, which results in tractable methods of learning the parameters and finding the optimal policy. The drawback of these models is again their simplicity: these models usually underfit and underestimate the actual environment behavior. For example, in robotics, the agent usually has noisy observations of the environment inner state and MDP is not a suitable model.

More complex models like Partially Observable Markov Decision Process (POMDP) can compensate for this drawback. Fitting this model to the environment, where the partial observation is given to the agent, generally gives dramatic performance improvement, sometimes unbounded improvement, compared to MDP. In general, finding the optimal policy for the POMDP model is computationally intractable and fully non convex, even for the class of memoryless policies. The open problem is to come up with a method to find an exact or an approximate optimal stochastic memoryless policy for POMDP models.
\end{abstract}





\section{Introduction}\label{s:intro}
The concept of planning, as a part of decision theory, in the AI literature 
 has a long history. It is the bases for a variety of popular agent-environment interaction problems like Reinforcement Learning (RL). RL is an effective approach to solve the problem of sequential decision making under uncertainty. RL agents learn how to maximize long-term reward using a experience obtained by direct interaction with a stochastic environment~\citep{bertsekas1996neuro-dynamic,sutton1998introduction}. Since the environment is initially unknown, the agent has to balance between \textit{exploring} the environment to estimate its structure, and \textit{exploiting} the estimates to compute a policy that maximizes the long-term reward. As a result, designing a RL algorithm requires three different elements: \textbf{1)} an estimator for the environment's structure, \textbf{2)} a planning algorithm to compute the optimal policy of the estimated environment~\citep{lavalle2006planning}, and \textbf{3)} a strategy to make a trade off between exploration and exploitation to minimize the   \textit{regret}, i.e., the difference between the performance of the exact optimal policy and the rewards accumulated by the agent over time.

Most of RL literature assumes that the environment can be modeled as a Markov decision process (MDP), with a Markovian state evolution that  is fully observed. A number of exploration--exploitation strategies have been shown to have strong performance guarantees for MDPs, either in terms of regret or sample complexity \cite{auer2009near}. 
  However, the assumption of full observability of the state evolution is often violated in practice, and the agent may have only noisy observations of the true state of the environment (e.g., noisy sensors in robotics). In this case, it is more appropriate to use the partially-observable MDP (POMDP)~\citep{sondik1971the-optimal} model. 

Many challenges arise in designing RL algorithms for POMDPs. Unlike in MDPs, the estimation problem (element 1) involves identifying the parameters of a latent variable model (LVM). 
The planning problem (element 2), i.e., computing the optimal policy for a POMDP with known parameters, is PSPACE-complete~\citep{papadimitriou1987the-complexity}, and it requires solving an augmented MDP built on a continuous belief space (i.e., a distribution over the hidden state of the POMDP). Finally, integrating estimation and planning in an exploration--exploitation strategy (element 3) with guarantees is non-trivial and no no-regret strategies are currently known.\\
 Previous works \cite{ross2007bayes} and \cite{poupart2008model-based} present new active learning algorithms to estimate the belief state in a model-based Bayesian RL approach, where a distribution over possible MDPs is updated over time. The proposed algorithms are such that the Bayesian inference can be done at each step, a POMDP is sampled from the posterior and the corresponding optimal policy is executed. The regret bound and sample complexity are not provided. 
 
Recently, the learning POMDP model parameter and imposing trade off between exploration and estimation (elements 1 and 3), are done at \cite{azizzadenesheli2016reinforcement}. They propose the theoretical guaranty on regret bound given the oracle memoryless policy. Therefore to close the learning, planing, and exploration-exploitation loop, the missing part, planning (element 2), is the remaining part. Therefore, planing  is a problem of finding the optimal memoryless policy, under uncertainty, in the class of stochastic memoryless polices. The overview complexity of planing in POMDP domain is discussed in \cite{kaelbling1998planning}.




\section{Formal Definition}\label{s:preliminaries}

\begin{figure}[t!]
\small
\begin{center}
\begin{psfrags}
\psfrag{x1}[][1]{$x_t$}
\psfrag{x2}[][1]{$x_{t+1}$}
\psfrag{x3}[][1]{$x_{t+2}$}
\psfrag{y1}[][1]{$\vec{y}_t$}
\psfrag{y2}[][1]{$\vec{y}_{t+1}$}
\psfrag{r1}[][1]{$\vec{r}_t$}
\psfrag{r2}[][1]{$\vec{r}_{t+1}$}
\psfrag{a1}[][1]{$a_{t}$}
\psfrag{a2}[][1]{$a_{t+1}$}
\includegraphics[width=0.6\textwidth,natwidth=810,natheight=642,trim={-1cm 0cm 0 11cm},clip]{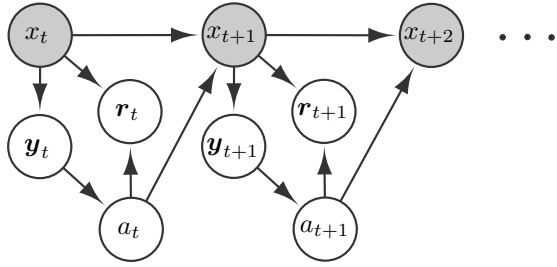}
\end{psfrags}
\end{center}
\vspace{-0.2in}
\caption{Graphical model of a POMDP under memoryless policies.}
\label{fig:pds2}
\end{figure}

A POMDP $M$ is a tuple $\langle \X, \A, \Y, \R, f_T, f_R, f_O\rangle$, where $\X$ is a finite state space with cardinality $|\X|=X$, $\A$ is a finite action space with cardinality $|\A|=A$, $\Y$ is a finite observation space with cardinality $|\Y|=Y$, and $\R$ is a finite reward space with cardinality $|\R|=R$ and largest reward $r_{\max}$.  In addition $f_T$ denotes the transition density, so that $f_T(x'|x,a)$ is the probability of transition to $x'$ given the state-action pair $(x,a)$, $f_R$ is the reward density, so that $f_R(\vec{r}|x,a)$ is the probability of receiving the reward in $\R$ corresponding to the value of the indicator vector $\vec{r}$ given the state-action pair $(x,a)$, and $f_O$ is the observation density, so that $f_O(\vec{y}|x)$ is the probability of receiving the observation in $\Y$ corresponding to the indicator vector $\vec{y}$ given the state $x$. Whenever convenient, we use tensor forms for the density functions such that $T_{i,j,l} = \Prob[x_{t+1}=j | x_t = i, a_t = l] = f_T(j|i,l),  s.t.~~T\in\Re^{X\times X\times A}$, $O_{n,i} = \Prob[\vec{y}=\vec{e}_n | x = i] = f_O(\vec{e}_n|i), s.t.~~ O\in\Re^{Y\times X}$, and $\Gamma_{i,l,m} = \Prob[\vec{r} = \vec{e}_m | x = i, a = l] = f_R(\vec{e}_m|i,l), s.t.~~\Gamma \in\Re^{X\times A\times R}.$
%
%
%
We also denote by $T_{:, j, l}$ the fiber (vector) in $\Re^{X}$ obtained by fixing the arrival state $j$ and action $l$ and by $T_{:,:,l}\in\Re^{X\times X}$ the transition matrix between states when using action $l$. The graphical model associated to the POMDP is illustrated in Fig.~\ref{fig:pds2}. 

We focus on stochastic memoryless policies which map observations to actions and for any policy $\pi$ we denote by $f_\pi(a|\vec{y})$ its density function. Acting according to a policy $\pi$ in a POMDP $M$ defines a Markov chain characterized by a transition density $f_{T,\pi}(x'|x) = \sum_a \sum_{\vec{y}} f_\pi(a|\vec{y}) f_O(\vec{y}|x) f_T(x'|x,a),$
%
%
and a stationary distribution $\omega_\pi$ over states such that $\omega_\pi(x) = \sum_{x'} f_{T,\pi}(x'|x)\omega_\pi(x')$. The expected average reward performance of a policy $\pi$ is $\eta(\pi; M) = \sum_x \omega_\pi(x) \wb{r}_\pi(x),$
%
%
where $\wb{r}_\pi(x)$ is the expected reward of executing policy $\pi$ in state $x$ defined as $\wb{r}_\pi(x) = \sum_a \sum_{\vec{y}} f_O(\vec{y}|x) f_\pi(a|\vec{y}) \wb{r}(x,a),$
%
and $\wb{r}(x,a) = \sum_{r} r f_R(r|x,a)$ is the expected reward for the state-action pair $(x,a)$.

The best stochastic memoryless policy is $\pi^* = \displaystyle\arg\max_{\pi} \eta(\pi; M)$ and we denote by $\eta^* = \eta(\pi^*; M)$ its average reward. Finding the optimal policy $\pi^*$ requires solving non-convex optimization and it is the desired open problem.



\section{Related Work}\label{ss:related}
Planning on uncertainty in a dynamic internal process is studied for infinite horizon \cite{sondik1978optimal}. 
 It is shown that, finding the exact optimal policy for POMDP is followed by the curse of dimensionality and the curse of history. People uses point-based value iteration \cite{pineau2006anytime} to reduce the complexity of the planning. It is also common to use heuristic search value iteration \cite{smith2004heuristic} and also policy tree with limited depth \cite{kaelbling1998planning} to reduce the planning complexity. 
For a finite horizon
But the computation complexity of finding optimal policy grows exponentially by horizon. For an infinite horizon, each vector of state distribution can be any point in the continues space of the simplex subspace. This means the planning is over the continuous space which is PSPACE-complete. \cite{sondik1978optimal} presented a method to partition the continuous space of the state distribution and then the policy is just a mapping from these partitions to the action.\\
In general, planning in the space of memoryless policy has a lower level of complexity. Although, it seems to be easier than belief based planning, it is still an $NP-hard$ problem \cite{vlassis2012computational}. To breaking down this complexity, \cite{li2011finding} presented a novel method for finding the optimal policy in the class of deterministic memoryless policies. Meanwhile, deterministic policies act poorly in the general case of POMDPs. The geometric representation of POMDP planning problem is shown in \cite{montufar2015geometry} and its geometric structure is well studied. Therefore, proposing a novel method to find the exact or approximated optimal memoryless policy (policy with performance $\epsilon-close$ to the performance of optimal policy) or limited history dependent policy under some mild conditions is the next step in the world of POMDP planning.




\small
\bibliography{Azizzadenesheli_open_problem_16}

\end{document}